\documentclass[10pt,twocolumn,letterpaper]{article}

\usepackage[pagenumbers]{iccv} 
\definecolor{iccvblue}{rgb}{0.21,0.49,0.74}
\definecolor{predictionpurple}{rgb}{0.6, 0, 1}
\definecolor{groundtruthblue}{rgb}{0, 0.6, 1}
\usepackage[pagebackref,breaklinks,colorlinks,allcolors=iccvblue]{hyperref}
\usepackage{graphicx}
\usepackage{amsmath}
\usepackage{placeins}
\usepackage{makecell}
\usepackage{algorithm}
\usepackage{algpseudocode}
\usepackage{float}
\usepackage{adjustbox}

\def\paperID{12863} 
\def\confName{ICCV}
\def\confYear{2025}

\title{ViewDelta: Scaling Scene Change Detection through Text-Conditioning}

\author{Subin Varghese\\
University of Houston\\
4226 MLK Blvd Houston TX \\
{\tt\small srvargh2@cougarnet.uh.edu}
\and
Joshua Gao\\
University of Houston\\
4226 MLK Blvd Houston TX \\
{\tt\small jkgao@cougarnet.uh.edu}
\and
Vedhus Hoskere\\
University of Houston\\
4226 MLK Blvd Houston TX \\
{\tt\small vhoskere@central.uh.edu}
}

\begin{document}
\maketitle
\begin{abstract}
We introduce a generalized framework for Scene Change Detection (SCD) that addresses the core ambiguity of distinguishing ``relevant'' from ``nuisance'' changes, enabling effective joint training of a single model across diverse domains and applications. Existing methods struggle to generalize due to differences in dataset labeling, where changes such as vegetation growth or lane marking alterations may be labeled as relevant in one dataset and irrelevant in another. To resolve this ambiguity, we propose \emph{ViewDelta}, a text conditioned change detection framework that uses natural language prompts to define relevant changes precisely, such as a single attribute, a specific set of classes, or all observable differences. To facilitate training in this paradigm, we release the Conditional Change Segmentation dataset (\emph{CSeg}), the first large-scale synthetic dataset for text conditioned SCD, consisting of over 500,000 image pairs with more than 300,000 unique textual prompts describing relevant changes. Experiments demonstrate that a single ViewDelta model trained jointly on CSeg, SYSU-CD, PSCD, VL-CMU-CD, and their unaligned variants achieves performance competitive with or superior to dataset specific models, highlighting text conditioning as a powerful approach for generalizable SCD. Our code and dataset are available at \href{https://joshuakgao.github.io/viewdelta/}{github.io/viewdelta/}.

\end{abstract}
    
\section{Introduction}
\label{sec:intro}

\begin{figure}
\centering
\includegraphics[width=0.45\textwidth]{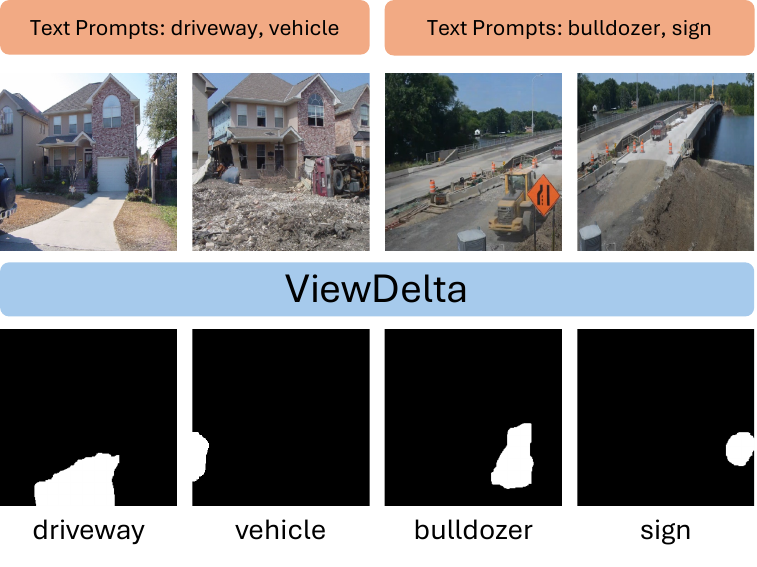}
\caption{Given a pair of (un)aligned images with text to contextualize what change may be of relevance, ViewDelta highlights  \textit{relevant} changes or all changes.
}
\label{fig:abstract_figure}
\end{figure}

Detecting and interpreting scene changes is a long–standing task in computer vision \cite{remotesenseingsurvey,singh1989review,jiang2022survey,cheng2024change}, with applications in situational awareness \cite{survelince1,survelince2}, disaster assessment \cite{saito2004using,disaster1,disaster2,disaster3}, and environmental monitoring \cite{env1,env2}.  
Although many public benchmarks exist, each encodes its own notion of what they label as a \emph{relevant} change. SYSU-CD assumes land-use changes in satellite imagery \cite{sysu_cd}, PSCD provides eight semantic street classes \cite{pscd}, and VL-CMU-CD focuses on changes such as cars, people, and litter \cite{vlcmucd}. 

Models trained on any single benchmark therefore learn a narrow, implicit definition of change and degrade sharply when deployed elsewhere \cite{Kim2024TowardsGS,Lin2024RobustSC}. Additionally, jointly training on different change detection datasets can be an ambiguous task for a model, as context is needed to understand what change is considered relevant. For example, if a model is given an image pair from a dataset that labels vegetation changes and another that ignores them, it receives contradictory signals about what features define a valid change. Resolving the ambiguity of what change is relevant is key to unlocking joint training across different sources, enabling the creation of a single model that can learn from a greater diversity of data, including from sources that have unaligned views between image pairs.

To address this ambiguity and investigate the performance of a jointly trained change detection model, we introduce the ViewDelta framework, \cref{fig:abstract_figure}, that incorporates text prompts to \emph{contextualize relevant change} for unaligned image pairs. This approach makes it feasible to unambiguously train a single model across multiple change detection datasets that define relevant change differently, thereby producing a more generalizable model.

In order to evaluate the ability of ViewDelta in distinguishing relevant and irrelevant change with varying text, we introduce a new dataset, CSeg (Conditional change Segmentation), consisting of over 500,000 images and 300,000 accompanying unique prompts. CSeg uses inpainting to create synthetic changes, similar in method to COCO-Inpainted \cite{Sachdeva2022TheCY,sachdeva2023change} and Changen2 \cite{sytheticdatagensatelite}, to simulate realistic change scenarios. \textbf{CSeg intentionally incorporates nuisance changes into images as well as irrelevant text in prompts}, thus requiring comprehension of the accompanying prompt to contextualize what is a relevant change.

ViewDelta is jointly trained on CSeg, PSCD \cite{pscd}, SYSU-CD \cite{sysu_cd}, VL-CMU-CD \cite{vlcmucd}, and the unaligned variants of PSCD and VL-CMU-CD \cite{Lin2024RobustSC}. We compare against fully finetuned models on each dataset and achieve competitive performance relative to domain-specific models across the datasets. Additionally, we evaluated what performance gain could be achieved by finetuning the generalized ViewDelta model for each dataset, showing a consistent improvement in performance. Our contributions include:

\begin{itemize}

\item Introducing a novel text prompt conditioned change detection task that outputs binary segmentations based on user-specified textual descriptions of relevant changes.

\item Proposing ViewDelta, a framework which enables joint training of change detection datasets that leverages prompt conditioning to disambiguate what change is relevant.

\item Creating and publicly releasing the CSeg dataset, consisting of aligned and unaligned image pairs with corresponding textual prompts and annotated segmentation masks.

\item Evaluation of the first generalized change detection model capable of performing scene change detection across varying views with user definable definition of relevant change or all changes at runtime across domains.

\end{itemize}
\section{Related Work}
\label{sec:related_work}

\textbf{Scene Change Detection.} Change detection (CD) for images is the umbrella task of identifying variations between two images of the same place taken at different times. CD methods have most commonly been applied in applications such as surveillance and satellite imagery \cite{review1,review2,review3,review4,sysu_cd}, where alignment between images is a reasonable assumption. CD has found widespread use in satellite imagery for both binary \cite{sysu_cd,levir_cd} and multi-class \cite{SECOND, Gupta2019xBDAD} change detection. 


Scene Change Detection (SCD), the primary focus of this work, narrows CD to ground‑level scenes where additional variations not usually accounted for in satellite images are faced. Example variations include in viewpoint, scale, and illumination \cite{pscd,vlcmucd,Sachdeva2022TheCY,Sachdeva2022TheCY}. Additionally, acquiring and labeling data for SCD is extremely laborious and introduces additional consideration to what constitutes as \textit{relevant} change \cite{pscd,vlcmucd,Wang2022HowTR}. For instance, whether shifting shadows, growing foliage, trash, structural degradation, or snow should be considered as a \emph{relevant} or \emph{nuisance} change. The designation of what is a relevant change depends entirely on the application. SCD models therefore learn what change is \emph{relevant} implicitly through the labels used for a dataset. The PSCD \cite{pscd} dataset introduced the first semantic SCD dataset with one of eight street-scene categories \emph{structure}, \emph{lane marking}, \emph{object (traffic)}, \emph{object (other)}, \emph{human}, \emph{vehicle}, and \emph{barrier}. Semantic SCD allows for explicit definition of \textit{relevant} change. Our work seeks to generalize SCD further by enabling a user to explicitly define \emph{relevant} change through text at runtime.

\textbf{Limitations of Foundational Segmentation Models.}
General purpose semantic segmentation approaches have had tremendous success with models such as the Segment Anything Model (SAM) \cite{kirillov2023segment, ravi2024sam2segmentimages}. Combined with image registration approaches, general semantic segmentation can provide valuable information for change detection. However, as shown in prior works in SCD \cite{pscd, Wang2022HowTR}, even a theoretically perfect open vocabulary semantic segmentation model struggles fundamentally with change detection in two common scenarios if (i) the object type remains the same (e.g., brick building) but changes still occur within an object (e.g., missing brick or new graffiti) and (ii) an object is replaced by another object of the same class (e.g., red car to blue car).

\textbf{Generalization in Change Detection.}
Recent works have increasingly focused on improving the generalization of SCD models to operate across SCD datasets \cite{Kim2024TowardsGS}, as well as operate on larger viewpoint variation in real images \cite{Lin2024RobustSC}. GeSCF \cite{Kim2024TowardsGS} introduced a framework for generalizable SCD, that utilizes SAM \cite{kirillov2023segment}, showing both a strong performance across datasets and nearly matches the performance of tailored SCD models like C-3PO \cite{Wang2022HowTR}. Impressively, GeSCF \cite{Kim2024TowardsGS} results also indicate better generalizability than finetuned SAM \cite{kirillov2023segment} based CD methods in terms of generalizability. However, GeSCF does not natively provide semantics for the changes found and requires coarse alignment between images.

Towards creating a generalizable model robust to view change and generalize across datasets, Lin et al. \cite{Lin2024RobustSC} proposed using the Dinov2 \cite{oquab2024dinov2learningrobustvisual} model with cross-attention, which we shall refer to as the Dinov2 RSCD model. To evaluate the performance of Dinov2 RSCD, Lin et al. \cite{Lin2024RobustSC} created two variants of both PSCD \cite{pscd} (Diff-1 PSCD and Diff-2 PSCD)\cite{oquab2024dinov2learningrobustvisual} and VL-CMU-CD \cite{vlcmucd} (VL-CMU-CD Diff-1 and VL-CMU-CD Diff-2 \cite{oquab2024dinov2learningrobustvisual}). These variants were created by perturbing the coarsely aligned image sequences to the first or second-nearest neighbor, thus allowing for true parallax and occlusion effects. Evaluation of Dinov2 RSCD showed state-of-the-art performance on both VL-CMU-CD, PSCD, and their variants. Our work closely aligns with Dinov2 RSCD \cite{Lin2024RobustSC} in that we also generalize across SCD datasets with view variations as well as leverage Dinov2 features in ViewDelta. However, we deviate in training methodology and further generalize the SCD task with text conditioning.

Large vision-language models now offer strong multimodal reasoning. Gemini 2.5 Pro can accept two images with a natural-language prompt and returns dense masks or bounding boxes. This interface lets us query it for text-conditioned change detection, aligning exactly with ViewDelta’s output format. We therefore include Gemini 2.5 Pro as a baseline and evaluate it on CSeg.

\textbf{Synthetic Dataset Generation.}
Creating SCD datasets of real images is an extremely laborious and expensive task \cite{pscd, vlcmucd}. To address this, several works have explored synthetic data generation. For instance, the COCO-Inpainted dataset \cite{Sachdeva2022TheCY} simulates change by inpainting and applying affine transformations to an image. In a follow-up study, KC-3D generates purely synthetic 3D scenes by randomly placing objects and progressively removing them across frames with varying camera viewpoints to create change pairs. More recently, Changen2 \cite{sytheticdatagensatelite}, leverages diffusion transformer models to simulate temporal changes by editing, removing, or generating object masks.
\section{ViewDelta}
\label{sec:architecture}
\begin{figure}
  \centering
  \includegraphics[width=0.45\textwidth]{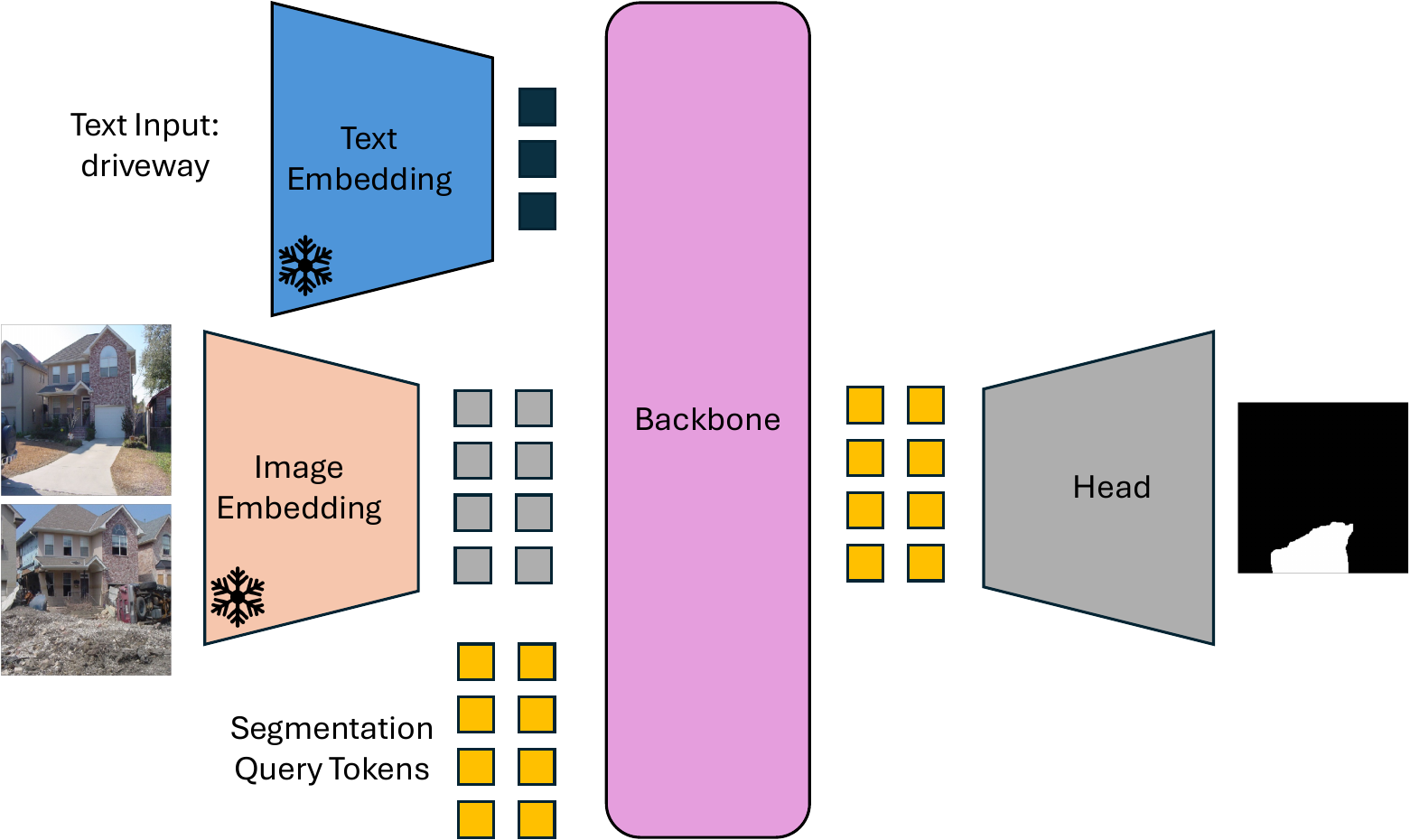}
  \caption{Overview diagram of the ViewDelta change detection framework, consisting of two primary branches to process images and text. The framework is agnostic to choice of embeddings, backbone, and segmentation head. We utilize SigLip text embeddings, Dinov2 image embeddings, a ViT backbone, and a fully convolutional mask decoder. The focus of this framework is to enable the evaluation of text prompts to enable joint training across SCD datasets.}
  \label{fig:viewdelta}
\end{figure}

In this section, we introduce \textbf{ViewDelta}, our proposed framework for text-prompted change detection, as illustrated in \cref{fig:viewdelta}. ViewDelta processes two input images, \( I_a \) and \( I_b \), captured before and after a change event, respectively, along with a text prompt \( T \) that specifies the type of change relevant to the user. The goal is to predict a binary segmentation map \( M \in \{0, 1\}^{H \times W} \), where \( M_{i,j} = 1 \) indicates the presence of the specified change at pixel location \( (i, j) \) and \( M_{i,j} = 0 \) otherwise. 

The proposed framework requires the adaptation of a multimodal model to take two images as well as a text prompt as input and output a binary segmentation mask for SCD. This deviates from existing SCD models as well as multi-modal segmentation models, as two images and text are used as input. From the proven effectiveness of transformer-based models in multi-modal applications \cite{clip,evaclip,zhai2023sigmoid}, we also utilize the Vision Transformer (ViT) \cite{Vit}, however we incorporate additional modifications to effectively integrate text prompts and generalizability with joint training for SCD. These modifications include integrating a frozen text encoder to transform the text prompt \(T\) into a sequence of tokens, using an image encoder instead of a traditional patch embedding network for processing the input images \(I_a\) and \(I_b\), and introducing learnable segmentation query tokens. We employ a conventional fully convolutional segmentation head to produce the final binary segmentation map \(M\) from the learnable segmentation query tokens, rather than directly from the image tokens. These design choices, as shown in our ablation, are required to achieve strong generalizability in SCD.

\subsection{Text Embeddings}
We leverage only the text encoder of the SigLip \cite{zhai2023sigmoid} model due to its improved performance against models such as CLIP \cite{clip}, OpenCLIP \cite{openclip}, and EVA-CLIP \cite{evaclip}. We leave the SigLip model frozen in order to keep the text generalization ability of SigLip. We evaluate the generalizability of the model on the CSeg dataset, which contains unique prompts in the test set that are not seen in the training set.

\subsection{Image Embeddings}
Patch embeddings have been widely adopted by state-of-the-art methods in both change detection and semantic segmentation. However, in our joint training configuration, we observed that optimizing patch embeddings was challenging due to the slow convergence rate on the available training data. We investigate directly using image features from a frozen Dinov2 \cite{oquab2024dinov2learningrobustvisual} model as embeddings. Our ablation study demonstrates that Dinov2 embeddings yielded faster convergence rate with less data. We attribute the difficulty of optimizing learned patch embeddings primarily to dataset size limitations rather than inherent issues with patch embeddings themselves.

\subsection{Backbone}
We utilize the ViT \cite{Vit} transformer architecture in its base configuration as the backbone, consisting of 12 transformer encoder layers, each characterized by a hidden dimension of 768, an intermediate MLP dimension of 3072, and 12 attention heads. While all image and text tokens are fed into the backbone, we do not employ feature aggregation strategies on the output tokens. Instead, we only pass the segmentation query tokens to the segmentation head. This differs from common aggregation strategies, such as addition, subtraction or concatenation, commonly employed in models like FC-EF \cite{FC_EF}, STANet \cite{STANet}, \cite{changemamba}, Dinov2 RSCD\cite{Lin2024RobustSC}, and DSAMNet \cite{sysu_cd}. As will be shown in our ablation, aggregating features during joint training showed suboptimal performance for our network configuration.

\subsection{Segmentation Query Tokens}
Drawing inspiration from DETR \cite{detr}, we leverage dedicated segmentation query tokens (SQT) to enhance model flexibility. These tokens enable the model to create streamlined representations for segmentation-related features, circumventing the inherent complexities encountered if the model had to directly map and align combined image and textual tokens to a feature space suitable for change detection. This methodological choice effectively addresses issues related to adding padding tokens or deciding which tokens from the sequence of image and text tokens to use as input to the segmentation head. As will be shown in our ablation section, removal of these segmentation query tokens causes a significant performance decrease.

\subsection{Segmentation Head}
Existing state-of-the-art change detection architectures, including SwinSUNet \cite{swinsunet}, ChangeFormerV4 \cite{changeformer}, and MambaBCD-Base \cite{changemamba}, predominantly assume spatial alignment between input image pairs, which introduces a bias that may limit performance in scenarios involving unaligned or varying viewpoints. To remove this bias, our segmentation head operates only on segmentation query tokens, as shown in \cref{alg:forward_prediction}. This design choice removes the necessity of explicit feature combination from spatially mismatched input images and facilitates effective fusion of visual and textual embeddings. The segmentation head itself is minimal in design consisting of a simple five-layer network following a sequence of a 2D convolution, a 2D transposed convolution, a second 2D convolution, bilinear upsampling, and a final 2D convolution, with the convolutional layers using ReLU activations.
\begin{algorithm}
\caption{Forward prediction of proposed model}
\label{alg:forward_prediction}
\begin{algorithmic}[1]
\Require Images $I_a$, $I_b$ and text embeddings $T_e$
\State $I_a \leftarrow \text{ImageEmbedder}(I_a)$
\State $I_b \leftarrow \text{ImageEmbedder}(I_b)$
\State $T \leftarrow \text{TextEmbedder}(T_e)$
\State $X \leftarrow \text{Concat}(I_a, I_b, T, \text{SQT})$
\State $X \leftarrow X + \text{PositionalEmbedding}$
\State $X \leftarrow \text{MLP}(X)$
\State $X \leftarrow \text{ViT}(X)$
\State $\text{SQT} \leftarrow X[:, -N_s:, :]$
\State $\text{SegLogits} \leftarrow \text{UpsampleNetwork}(\text{SQT})$
\State \Return $\text{SegLogits}$
\end{algorithmic}
\end{algorithm}

\subsection{Training}
\textbf{Joint Dataset.} ViewDelta is trained jointly on CSeg, SYSU-CD, PSCD (both its binary and multi-class forms), VL-CMU-CD, plus the Diff-1 and Diff-2 variants of PSCD and VL-CMU-CD. For PSCD, each semantic label is turned into a separate binary mask, and the class name becomes the text prompt. Image pairs that contain no changes for the given prompt are kept in both training and testing to ensure fair comparison with existing multi-class SCD methods.

\textbf{Training Parameters.} During training, the Adam optimizer \cite{kingma2014adam} with weight decay, a one-epoch warm-up, and cosine annealing learning rate decay, starting from an initial learning rate of \(2 \times 10^{-5}\) is used. Training is performed using a batch size of 4 per GPU using 4 Nvidia A100 GPUs, resulting in an effective batch size of 16. We utilize DeepSpeed ZeRO Stage 2 \cite{deepspeed} for sharding optimizer states and gradients, activation checkpointing within attention layers and large linear layer, and 16-bit mixed-precision training.
\section{CSeg Dataset}

\begin{figure}
  \centering
  \includegraphics[width=0.46\textwidth]{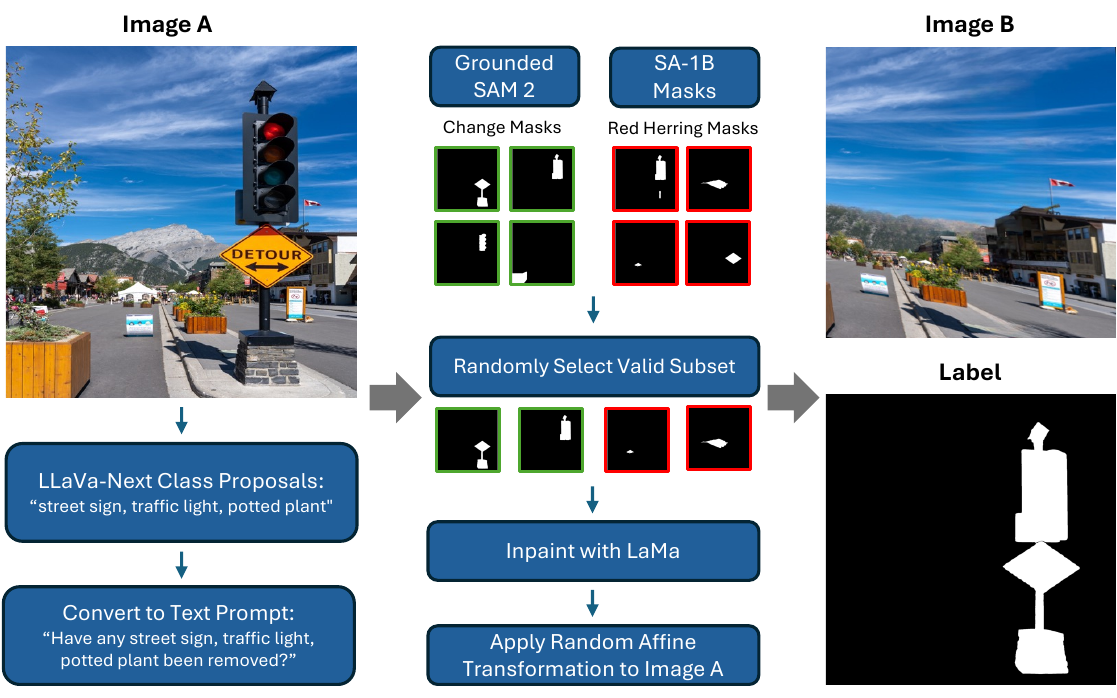}
  \caption{\textbf{Outline of the steps involved in generating CSeg.} LLaVA-Next identifies object classes in an SA-1B image, which inform a text prompt and guide Grounded SAM 2 in extracting instance masks. Some masks are then inpainted to simulate changes, while some original SA-1B masks are used as ``red herrings". This process yields a text prompt, Image A, Image B, and a change mask label.}
  \label{fig:cseg_pipeline}
\end{figure}


Inspired by the COCO-Inpainted dataset \cite{sachdeva2023change}, we propose a procedure as shown in \cref{fig:cseg_pipeline} that leverages state-of-the-art vision models to generate the CSeg dataset. CSeg is a large-scale, synthetic dataset designed to evaluate a model's ability to distinguish relevant from nuisance changes in images based on a given text prompt to contextualize what may be relevant. 

\subsection{Generating Changes and Text Prompts}

Similar to the COCO-Inpainted dataset \cite{sachdeva2023change}, we simulate changes between two images by inpainting and applying an affine transformation for view variation. However, our approach is unique in its use of large vision-language models to identify objects to  generate diverse text prompts. We first select an SA-1B \cite{kirillov2023segment} image and leverage LLaVa-Next \cite{liu2024llavanext} to identify a maximum of ten different objects in the image to be used as classes. To promote dataset balance, we select one to five of the least represented classes during generation. These classes are then used directly as text prompts or applied to a natural language prompt template. These prompt templates are a list of 45 predefined and manually validated templates generated with GPT-4o that have the intent of identifying change. To simulate change, Grounded SAM 2 \cite{ravi2024sam2segmentimages, liu2023grounding, ren2024grounded, ren2024grounding, kirillov2023segment, jiang2024trex2} is then used to generate instance masks across classes, and we randomly select a subset of at most 10 masks to be inpainted in Image B with LaMa \cite{suvorov2021resolution}. We combine these inpainted masks to produce the change label. Finally, we apply a random affine transformation to either Image A or Image B to simulate a perspective change. To account for potential occlusion caused by the affine transformation, we check the masks of the inpainted objects and remove any occluded masks from the change label.

ViewDelta should be generalized to detect all changes in an image and not limited to those specified by a prompt, which eliminates the need for explicit class enumeration. To generate these ``all" image pairs, we randomly inpaint five to ten SA-1B masks on Image B to simulate change. The text prompt is randomly selected from a predefined list of 96 prompts generated with GPT-4o that convey an intent for all changes.

\subsection{Combating Inpainting Noise}

As stated in \cite{sachdeva2023change}, the inpainted regions tend to have inpainting artifacts (also observed in other studies \cite{cheng2024change, jia2003image, criminisi2004region, 9879161}). We follow a similar strategy employed in the COCO-Inpainted dataset \cite{sachdeva2023change} where we include ``red herring" masks to prevent the model from learning artifacts instead of image changes. These red herring masks are a randomly selected subset of zero to ten SA-1B masks that are not classified as any of the classes in the text prompt. Since SA-1B doesn't provide mask classifications, we use Grounded SAM 2 with a low confidence threshold to ensure that the SA-1B mask doesn't coincide with a change class instance. These red herring masks are inpainted on image B, but not added to the change mask. During training, this forces the model to learn the change between the two images that are consistent with the text prompt, rather than labeling inpainting noise as changes. Note that adding red herring masks to ``all" image pairs is not applicable, since every change needs to be accounted for, which would include all inpaints.

\subsection{CSeg Dataset Statistics}

\begin{figure}
  \centering
  \includegraphics[width=.47\textwidth]{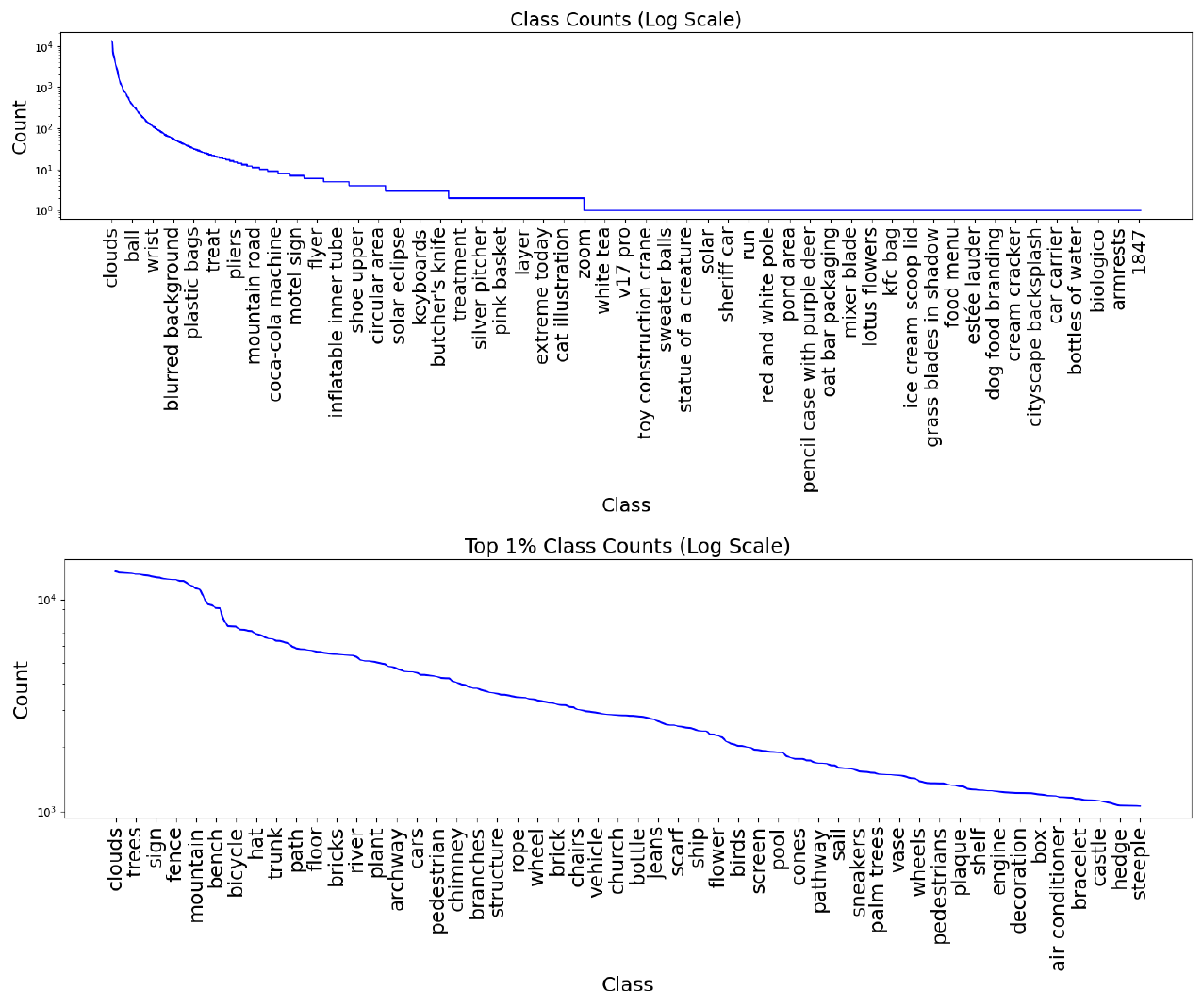}
  \caption{\textbf{Class count distributions.} We separate the top 1\% most frequent classes and the complete dataset for easier evaluation. These figures don't include the ``all" class.}
  \label{fig:cseg_class_counts}
\end{figure}

There is a total of 501,153 change image pairs in CSeg split into 451,090 pairs in train and 50,063 in test. There are 24,298 unique classes and 339,348 unique prompts in train and 7,285 unique classes and 7,326 unique prompts in test. There are 1,408 unique classes and 35,271 unique prompts not seen in train. 53,257 and 5,947 pairs were ``all" image pairs in the train and test sets respectively at about a 12\% of the total for each set. CSeg's class counts are top-heavy in that more common classes such as ``clouds", ``sky", ``people", ``water", ``trees" and ``roof" appear exponentially more often than less common classes like ``calcite", ``aldravanda", and ``ice hammer" as depicted in \cref{fig:cseg_class_counts}. This is due to how the majority of images in SA-1B include these common objects. During class proposals, we select the classes that are least used in CSeg to mitigate this problem.

\begin{figure}
  \centering
  \includegraphics[width=.45\textwidth]{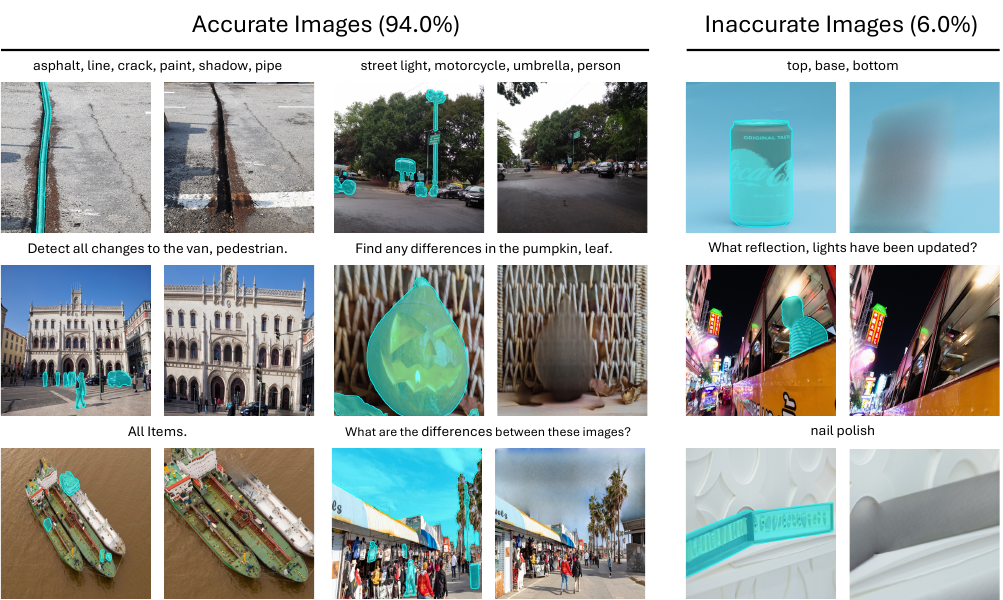}
  \caption{\textbf{Evaluation of CSeg label quality. }Change mask labels are highlighted in blue. Manual inspection of the generated images show a 94\% agreement with a human reviewer.}
  \label{fig:cseg_qualitative}
\end{figure}

Although LLaVa-Next \cite{liu2024llavanext} and Grounded SAM 2 \cite{ravi2024sam2segmentimages, liu2023grounding, ren2024grounded, ren2024grounding, kirillov2023segment, jiang2024trex2} are very impressive vision models, they cannot produce flawless class proposals and instance masks for all images across all classes. This consequently affects the quality of CSeg. To quantitatively measure the quality of CSeg, we manually check randomly sampled images for change label correctness until the percent accuracy clearly converges. We validated 500 images - about 0.1\% of the entire dataset - and observed an accuracy of 94.0\%. This results in a margin of error of ±2.22\% at a 95\% statistical confidence. The large majority of errors stem from Grounded SAM 2 yielding masks that are misclassified, especially with more challenging classes. We provide additional examples in \cref{fig:cseg_qualitative} for further evaluation.

\section{Experiments}
\label{sec:experiments}



\begin{figure*}[!h]
  \centering
  \begin{subfigure}[t]{0.6\textwidth}
    \centering
    \includegraphics[width=\textwidth]{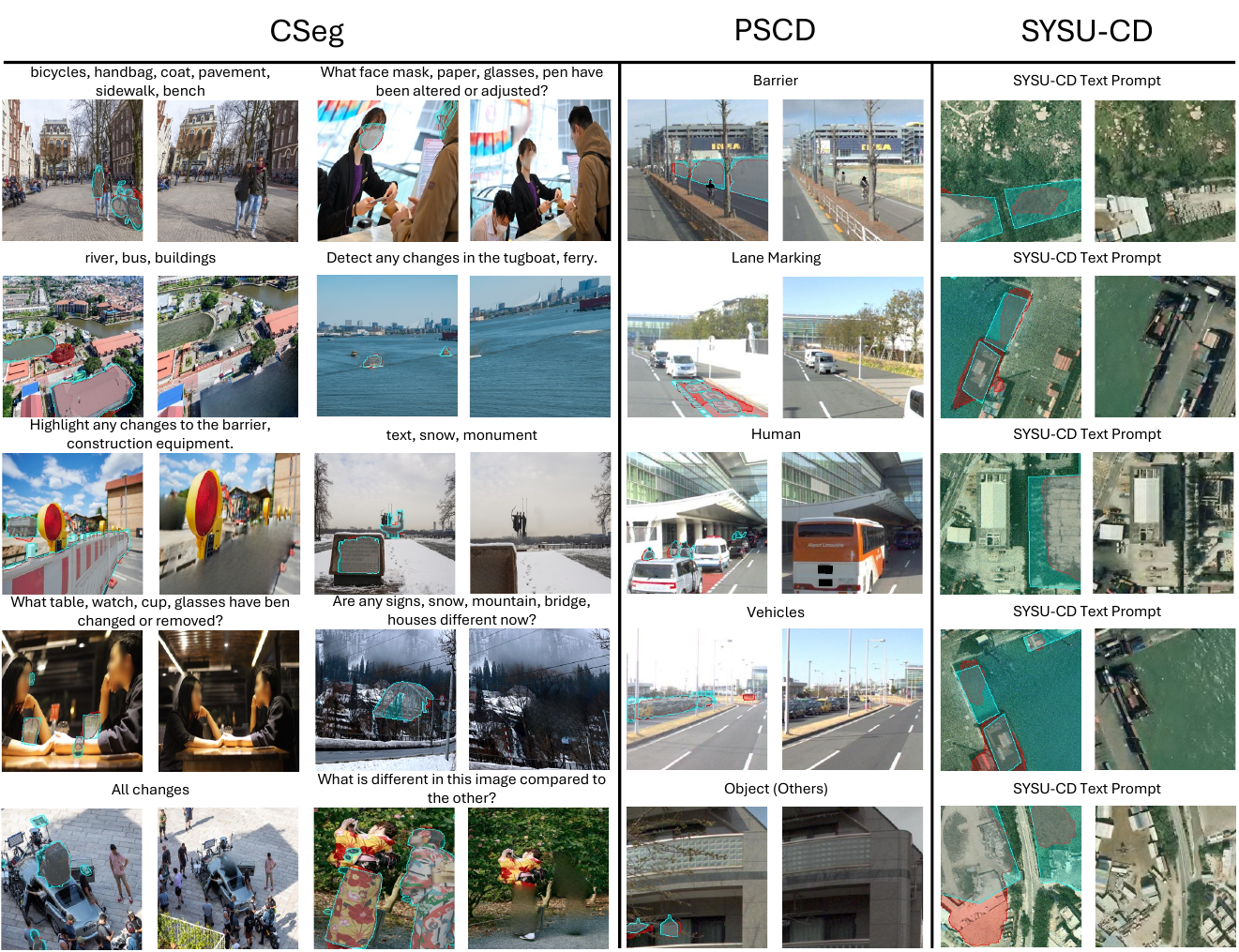}
    \caption{\textbf{Qualitative Results}: We show predictions from the jointly trained ViewDelta model in red, along with the ground truth in blue. \textbf{The SYSU-CD Text Prompt is} ``urban development, suburban expansion, pre-construction groundwork, vegetation alteration, road widening, and coastal construction".}
    \label{fig:qualitiative_evaluation}
  \end{subfigure}
  \hfill 
  \begin{subfigure}[t]{0.3\textwidth}
    \centering
    \includegraphics[width=0.9\textwidth]{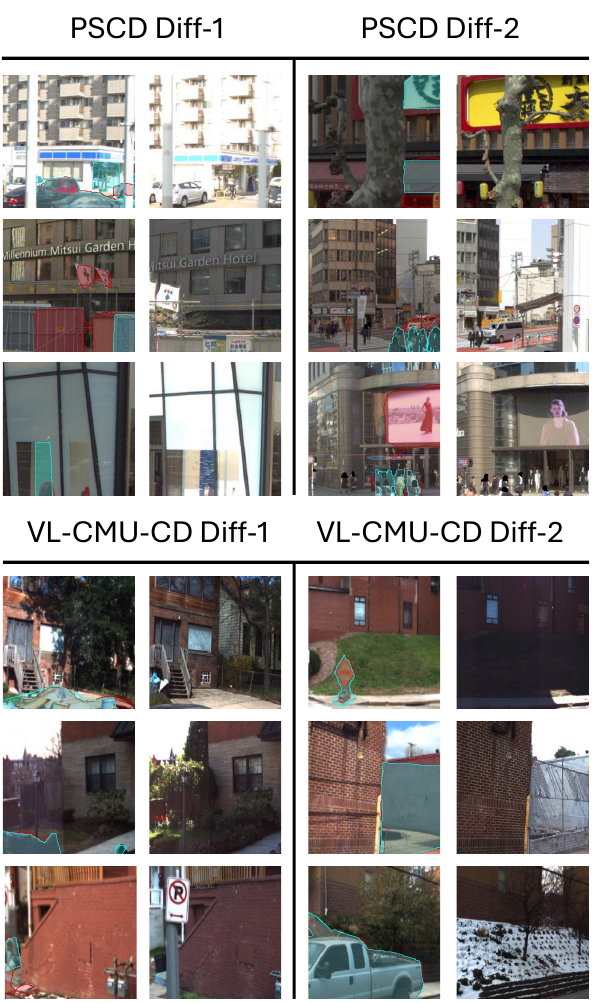}
    \caption{\textbf{Qualitative Results on unaligned image pairs.} Jointly trained ViewDelta model predictions (red) and ground truth labels (blue).}
    \label{fig:qualitiative_evaluation_unaligned}
  \end{subfigure}
  
  \caption{Qualitative evaluation results showing model predictions across different scenarios.}
  \label{fig:combined_qualitative}
\end{figure*}

We evaluate ViewDelta in two configurations: (1) a general model jointly trained across multiple datasets (CSeg, PSCD, VL-CMU-CD, SYSU-CD, and their unaligned variants), and (2) specialized models that start from the general model and are then fine-tuned on individual datasets. To assess effectiveness across diverse scenarios, we test on five benchmarks: CSeg (our proposed dataset with complex text prompts), PSCD \cite{pscd} (street-view semantic changes), SYSU-CD \cite{sysu_cd} (satellite imagery), VL-CMU-CD \cite{vlcmucd} (street-view with label noise), and the Diff-1/Diff-2 unaligned variants of PSCD and VL-CMU-CD \cite{Lin2024RobustSC} (testing robustness to viewpoint changes). These datasets span indoor, outdoor, street-view, and satellite imagery domains.

We report standard change detection metrics including Intersection over Union (IoU), F1 score, Recall, and Precision. For each dataset, we adopt task-appropriate prompting strategies: dataset-specific class names for PSCD, a comprehensive change description for SYSU-CD, and a description of object categories for VL-CMU-CD. Section \ref{sec:experiments} presents quantitative comparisons with state-of-the-art methods, qualitative analysis of model predictions, and ablation studies examining key architectural components.

\subsection{Quantitative Evaluation}

\begin{table}
    \begin{adjustbox}{max width=\columnwidth}
        \setlength{\tabcolsep}{3pt}  
        \footnotesize               
    \centering
    \begin{tabular}{lcccc}
        \toprule
        \textbf{Model} & \textbf{IoU} & \textbf{F1} & \textbf{Rec} & \textbf{Prec} \\
        \midrule
        ViewDelta  & 83.80 & 91.19 & 87.61 & 95.07 \\
        Gemini 2.5 & 37.15 ± 1.48 & 54.18 ± 1.63 & 58.31 ± 1.72 & 50.59 ± 1.80 \\
        \bottomrule
    \end{tabular}
    \end{adjustbox}
    \caption{\textbf{Quantitative Results on the CSeg Dataset.} The table compares ViewDelta, evaluated on the full test set, with a baseline from Gemini 2.5 Pro. Due to the nature of API-based model access, Gemini's performance was assessed on a randomly selected subset of 2000 test samples. We report the standard error at a 95\% statistical confidence for these results to reflect the statistical variance inherent in this sampling approach.}
    \label{tab:cseg_results}
\end{table}

\begin{table}
    \centering
    \begin{tabular}{lcccc}
        \toprule
        \textbf{Model} & \textbf{IoU} & \textbf{F1} & \textbf{Rec} & \textbf{Prec}  \\ 
        \midrule
        \multicolumn{5}{c}{\textit{Fine-tuned models}}\\
        \midrule
        \makecell{CSCDNet \\ + SSCDNet \cite{pscd}} & 22.3 & -- & -- & -- \\
        CSSCDNet \cite{pscd} & 32.2 & -- & -- & -- \\
        ViewDelta   & \textbf{55.5} & \textbf{78.3} & 69.5 & \textbf{73.5} \\
        \midrule
        \multicolumn{5}{c}{\textit{General model}}\\
        \midrule
        ViewDelta  & 51.2 & 67.8 & \textbf{75.2} & 61.7 \\
        \bottomrule
    \end{tabular}
\caption{\textbf{Quantitative Results for semantic SCD on the PSCD \cite{pscd} Dataset.} Scores are averaged over all classes. To evaluate our text-conditioned model on this multi-class benchmark, we prompt ViewDelta with each class name (e.g., ``human") for every image pair. This procedure is followed regardless of which changes are present, ensuring a fair comparison against standard multi-class SCD methods.}
    \label{tab:pscd_results}
\end{table}

\begin{table}
    \centering
    \begin{tabular}{lcccc}
        \toprule
        \textbf{Method} & \textbf{IoU} & \textbf{F1} & \textbf{Rec} & \textbf{Prec} \\
        \midrule
        \multicolumn{5}{c}{\textit{Fine-tuned models}}\\
        \midrule
        FC-EF \cite{FC_EF} & 61.04 & 75.81 & 75.17 & 76.47 \\
        FC-Siam-diff \cite{FC_EF} & 61.01 & 75.79 & 75.30 & 76.28 \\
        FC-Siam-conc \cite{FC_EF} & 60.23 & 75.18 & 76.75 & 73.67 \\
        BiDateNet \cite{BiDateNet} & 62.52 & 76.94 & 72.60 & 81.84 \\
        STANet \cite{STANet} & 63.09 & 77.37 & \textcolor{blue}{\textbf{85.33}} & 70.76 \\
        DSAMNet \cite{sysu_cd} & 64.18 & 78.18 & 81.86 & 74.81 \\
        ChangeFormerV4 \cite{changeformer} & 65.03 & 78.81 & 77.90 & 79.74 \\
        BIT-50 \cite{BiT} & 66.14 & 79.62 & 77.90 & 81.42 \\
        TransUNetCD \cite{transunedcd} & 66.79 & 80.09 & 77.73 & 82.59 \\
        SwinSUNet \cite{swinsunet} & 68.89 & 81.58 & 79.75 & 83.50 \\
        MambaBCD \cite{changemamba} & \textcolor{blue}{\textbf{71.10}} &\textcolor{blue}{\textbf{83.11}} & \textcolor{red}{\textbf{80.31}} & \textcolor{red}{\textbf{86.11}} \\
        ViewDelta & \textcolor{red}{\textbf{70.09}} & \textcolor{red}{\textbf{82.41}} & 79.91 & 85.08 \\
        \midrule
        \multicolumn{5}{c}{\textit{General model}}\\
        \midrule
        ViewDelta & 67.05 & 80.27 & 73.90 & \textcolor{blue}{\textbf{87.91}} \\
        \bottomrule
    \end{tabular}
    \caption{\textbf{Quantitative Results on SYSU-CD Dataset.} The highest values are highlighted for \textcolor{blue}{\textbf{First}} and \textcolor{red}{\textbf{Second}}. ViewDelta is given a constant prompt of ``urban development, suburban expansion, pre-construction groundwork, vegetation alteration, road widening, and coastal construction" to allow for a fair evaluation.}
    \label{tab:sysu_cd_results}
\end{table}

\begin{table}
    \centering
    \begin{tabular}{lccc}
        \toprule
        \textbf{Method} & \textbf{Aligned} & \textbf{Diff-1} & \textbf{Diff-2} \\
        \midrule
        \multicolumn{4}{c}{\textit{Fine-tuned models}}\\
\midrule
DR-TANet\cite{Chen2021DRTANetDR} & 19.0 & 16.9 & 12.5 \\
C-3PO\cite{Wang2022HowTR} & 43.3 & 24.6 & 16.5 \\
Dinov2 RSCD\cite{Lin2024RobustSC} & 44.2 & 28.4 & 19.1 \\
ViewDelta & \textbf{63.1} & \textbf{63.6} & \textbf{62.5} \\
\midrule
\multicolumn{4}{c}{\textit{General model}}\\
\midrule
ViewDelta & 56.2 & 58.6 & 62.2 \\
\bottomrule
    \end{tabular}
    \caption{\textbf{Quantitative Results (F1 scores) for binary change detection on the unaligned variants of PSCD \cite{Lin2024RobustSC,pscd}}. ViewDelta is given the prompt ``all" to allow for a fair comparison between methods.}
    \label{tab:pscd_finetune_f1}
\end{table}

\begin{table}
    \centering
    \begin{tabular}{lccc}
        \toprule
        \textbf{Method} & \textbf{Aligned} & \textbf{Diff-1} & \textbf{Diff-2} \\
        \midrule
        \multicolumn{4}{c}{\textit{Fine-tuned models}}\\
        \midrule
        DR-TANet\cite{Chen2021DRTANetDR} & 60.7 & 57.7 & 56.9 \\
        C-3PO\cite{Wang2022HowTR} & 79.5 & 72.1 & 69.3 \\
        Dinov2 RSCD\cite{Lin2024RobustSC} & 79.5 & 76.0 & 73.9 \\
        ViewDelta & 81.4 & \textbf{79.5} & \textbf{78.2} \\
        \midrule
        \multicolumn{4}{c}{\textit{General models}}\\
        \midrule
        GeSCF\cite{Kim2024TowardsGS} & 75.4 & - & - \\
        ViewDelta & \textbf{81.8} & 78.9 & 76.1 \\
        \bottomrule
    \end{tabular}
    \caption{\textbf{Quantitative Results (F1 scores) for binary change detection on the unaligned variants of VL-CMU-CD \cite{Lin2024RobustSC,vlcmucd}.} Due to label noise \cite{Kim2024TowardsGS,Wang2022HowTR} in VL-CMU-CD, we use a single prompt ''Bins, Signs, Traffic-signs, Vehicles, Refuse, Construction, Maintenance Work, Buildings" as these are the changes we have found to be consistent in the human labels. The fine-tuned model's decrease relative to the general model reflects overfitting to noise.}
    \label{tab:vl_cmu_cd_finetune_f1}
\end{table}
We evaluate ViewDelta's core capabilities: (1) interpreting natural language to identify relevant changes, (2) generalizing across diverse scene types, and (3) handling viewpoint variations. Our experiments compare the general jointly-trained model against dataset-specific fine-tuned versions and state-of-the-art baselines.

\textbf{Text-Prompted Change Detection (CSeg).} Table~\ref{tab:cseg_results} demonstrates ViewDelta's ability to interpret complex natural language prompts on our proposed CSeg dataset. The general ViewDelta achieves 83.80\% IoU, significantly outperforming Gemini 2.5 Pro. The high precision (95.07\%) indicates ViewDelta effectively filters nuisance changes based on text specifications.

\textbf{Cross-Domain Generalization.} Tables~\ref{tab:pscd_results} and \ref{tab:sysu_cd_results} reveal ViewDelta's generalization across street-view (PSCD) and satellite (SYSU-CD) domains. On PSCD semantic SCD, our general model achieves 51.23\% IoU without any dataset-specific training, only 4.31\% below the fine-tuned version (55.54\%). This small performance gap demonstrates effective transfer learning from joint training. Notably, the general model achieves higher recall (75.16\% vs 69.47\%), suggesting it learns more inclusive change representations across datasets.

On SYSU-CD satellite imagery, the general ViewDelta achieves 67.05\% IoU, outperforming several methods but falling short from state of the art. The fine-tuned version reaches 70.09\% IoU, placing second only to MambaBCD \cite{changemamba}. Remarkably, the general model achieves the highest precision among all methods (87.91\%)

\textbf{Robustness to Viewpoint Changes.} Tables~\ref{tab:pscd_finetune_f1} and \ref{tab:vl_cmu_cd_finetune_f1} evaluate performance on unaligned image pairs with increasing viewpoint differences. On PSCD variants, ViewDelta shows minimal performance degradation. We attribute this robustness to our design choice to avoid spatial alignment assumptions in the segmentation head.




\subsection{Qualitative Evaluation}

We present qualitative results of our model on example image pairs from the CSeg, PSCD, and SYSU-CD test sets in \cref{fig:qualitiative_evaluation}. For fair comparison on the SYSU-CD test set, all images use the text prompt: ``urban development, suburban expansion, pre-construction groundwork, vegetation alteration, road widening, and coastal construction." For the PSCD test set, each image uses a prompt corresponding to the classes defined in the dataset, such as ``vehicle" or ``structure." Qualitative results are also shown for the unaligned variants of PSCD and VL-CMU-CD in \cref{fig:qualitiative_evaluation_unaligned}. Additional qualitative analysis is provided in the supplementary material presenting the effect of prompt variations across datasets, such as misspellings, semantic equivalents, and interrogative formulations. 


\subsection{Ablation Study}

We conduct comprehensive ablations to understand the contribution of each architectural component to ViewDelta's performance. Our analysis reveals critical design choices for effective text-prompted change detection.

\textbf{Choice of Embeddings.} Table~\ref{tab:cseg_subset_model_ablation} evaluates different combinations of text and image encoders on a subset of CSeg (22,624 images, 12-hour training). The SigLip + Dinov2 combination achieves 56.44\% IoU, more than doubling the performance of SigLip + Patch Embedding (26.16\%). This improvement demonstrates that pretrained visual features are crucial for data-efficient learning. Interestingly, SigLip consistently outperforms CLIP regardless of image encoder, validating our choice of text encoder for its superior vision-language alignment capabilities.

\textbf{Component Analysis.} Table~\ref{tab:ablation_study} examines the contribution of each architectural component through systematic removal.

\textit{1. Frozen Dinov2 features improve performance across domains.} Replacing Dinov2 with trainable patch embeddings reduces performance on all datasets, with PSCD showing the largest relative decrease and CSeg the smallest. 

\textit{2. Segmentation Query Tokens (SQT) are critical for multimodal fusion.} Removing SQT substantially reduces performance, especially on PSCD.

\textit{3. Text prompts significantly impact performance on semantic tasks.} SYSU-CD maintains reasonable performance, while performance drops substantially on CSeg and PSCD. This pattern suggests text prompts are most critical when distinguishing between multiple semantic categories, as in PSCD, versus binary change detection tasks like SYSU-CD.

\begin{table}[h]
    \centering
    \begin{tabular}{lc}
        \toprule
        \textbf{Embedding Configurations} & \textbf{IoU} \\
        \midrule
        SigLip + Dinov2 & \textbf{56.44} \\
        SigLip + Patch Embedding & 26.16 \\
        CLIP + Dinov2 & 53.46 \\
        CLIP + Patch Embedding & 27.56 \\
        \bottomrule
    \end{tabular}
    \caption{\textbf{Impact of varying embedding combinations on ViewDelta on a subset of CSeg.}}
    \label{tab:cseg_subset_model_ablation}
\end{table}



\begin{table}[h]
    \centering
    \begin{tabular}{lccc}
        \toprule
        \textbf{Dataset} & \textbf{CSeg} & \textbf{PSCD \cite{pscd}} & \textbf{SYSU-CD \cite{sysu_cd}} \\
        \midrule
        ViewDelta & \textbf{85.91} & \textbf{52.24} & \textbf{68.59} \\

        \midrule
        w/o Dinov2 & 61.05 & 16.87 & 34.81 \\
        w/o SQT & 62.48 & 8.43 & 53.55 \\
        w/o Prompts & 77.72 & 10.38 & 67.34 \\
        \bottomrule
    \end{tabular}
    \caption{\textbf{IoU of ViewDelta's main components when jointly trained on Cseg, PSCD, and SYSU-CD.} ``w/o Dinov2'': Replaces the frozen Dinov2 backbone with a trainable patch embedding. ``w/o SQT'': Removes the segmentation query tokens and generates the change mask from the image tokens. ``w/o Prompt": ViewDelta trained with no prompt.}
    \label{tab:ablation_study}
\end{table}


\subsection{Limitations}
Although CSeg supplies a diverse set of prompts for evaluating how well a model identifies relevant changes, its variations are largely synthetic. Though we evaluate the effects of real parallax and occlusions in PSCD and VL-CMU-CD, the prompt vocabularies are restricted to eight and one classes, respectively. To the best of our knowledge, no large scale real dataset currently exists with varying views and varying prompts to contextualize relevant change.
\section{Conclusion}
\label{sec:conclusion}
ViewDelta introduces text-conditioned scene change detection, enabling joint training across datasets with different labeling conventions by using natural language prompts to define relevant changes. The architecture leverages frozen pretrained embeddings and segmentation query tokens to handle viewpoint variations without spatial alignment assumptions. Experiments on our CSeg dataset (500K+ image pairs, 300K+ unique prompts) and existing benchmarks show that a single jointly-trained model achieves performance competitive with dataset-specific methods. Text conditioning proves effective for building generalizable change detection systems that adapt to user-specified relevance at inference time.
\FloatBarrier
{
    \small
    \bibliographystyle{ieeenat_fullname}
    \bibliography{main}
}

\end{document}